# A Comprehensive Study of Sparse Representation Techniques for Offline Signature Verification

E. N. Zois, D. Tsourounis, I. Theodorakopoulos, A. L. Kesidis and G. Economou

**Abstract**— In this work, a feature extraction method for offline signature verification is presented that harnesses the power of sparse representation in order to deliver state-of-the-art verification performance in several signature datasets like CEDAR, MCYT-75, GPDS and UTSIG. Beyond the accuracy improvements, several major parameters associated with sparse representation; such as selected configuration, dictionary size, sparsity level and positivity priors are investigated. Besides, it is evinced that $2^{nd}$ order statistics of the sparse codes is a powerful pooling function for the formation of the global signature descriptor. Also, a thorough evaluation of the effects of preprocessing is introduced by an automated algorithm in order to select the optimum thinning level. Finally, a segmentation strategy which employs a special form of spatial pyramid tailored to the problem of sparse representation is presented along with the enhancing of the produced descriptor on meaningful areas of the signature as emerged from the BRISK key-point detection mechanism. The obtained state-of-the-art results on the most challenging signature datasets provide a strong indication towards the benefits of learned features, even in writer dependent (WD) scenarios with a unique model for each writer and only a few available reference samples of him/her.

## 1 INTRODUCTION

ONE of the most acceptable and long-standing behavioral modus to declare and verify a person's identity or acknowledge of his/hers consent in a wide variety of cases is the handwritten signature. Signatures are depicted by their trace, usually onto a sheet of paper or an electronic device. This process conveys information related not only to the imprinting of the personal details of the signatory, but also to his/her writing system and psychophysical state [1]. Ongoing research regarding the development of offline or static Automated Signature Verification Systems (ASV's/SV's) indicate clearly that this topic still is an active, open and important research field [2-6].

The purpose of an offline SV system is to recognize an image of a signature in question as genuine or forgery, i.e. to verify the writer's genuine signatures and reject the forgery ones. This work addresses writer dependent signature verification (WD) by employing a unique model for each participating writer with the use of some genuine reference samples and some genuine samples from other writers as representatives of the positive and negative class respectively.

Conceivably, the most influential step in the design of a SV system is the feature extraction stage, which can be defined as the process that maps any given signature image into a multidimensional vector. Feature extraction methods can be divided into two major categories: a) handcrafted features which aim at pre-determined characteristics of the image; examples of this branch include methods with global-local and/or grid-texture oriented features [7] and b) features learned directly from images, where Deep Learning (DL) [2, 8-10], Bag of Visual Words (BoW) [11-13] or Histogram of Templates (HOT) [14] are some of the most representative techniques.

Learning features from images could potentially have significant advantages, since such techniques -in general- can discover specialized spatial associations that are inherent into the signature images. Among the most powerful categories of such methods is Sparse Representation (SR). At a glance, SR methods cope with the problem of signal representation. Frequently they involve an overcomplete basis or dictionary D which can be developed either by using a mathematical model or a set of data instances. The term "overcomplete" means that the population of the dictionary members (or atoms) is greater with respect to the dimensionality of one atom. Then, given a dictionary D along with an unknown signal X, SR handles the reconstruction, in an optimization context, of X by a linear combination of the *atoms*. The reconstruction is constrained by the fact that only as few as possible atoms are selected from the overcomplete dictionary for the minimization of the reconstruction error, something that promotes the sparsity of the representation. The benefit of having an over-complete basis is that the basis vectors can adapt to patterns inherent in the input data. Nevertheless, the use of an over-complete basis, may lead to coefficients which are no longer uniquely determined. Consequently, in SR, the additional criterion of sparsity is applied in order to resolve the degeneracy introduced by over-completeness. Figure 1, depicts the SR concept.

Sparse Representation techniques are the subject of scientific interest for quite a long time [15-18] and have been proved to be extremely useful in computer vision and pattern recognition applications. This work presents for the first time a method for off-line signature verification that harnesses the power of SR in order to deliver state-of-the-art verification performance with the use of few genuine reference samples. The current paper introduces several advancements in the line of work initiated with two recently published papers investigating the potential of SR [19] and Archetypal Analysis (AA) [20] on the offline SV task. Beyond the accuracy improvements and the high-end performance achieved in several popular signature datasets, this work has a significant contribution towards the following directions:

1. We innovatively address several aspects of local feature pooling for offline SV. As a result, we provide evidence of the superiority of 2nd order statistics of the sparse coefficients as a powerful pooling function for the formation of the global signature descriptor. Combined with a segmentation strategy which employs a special form of spatial pyramid (SP) tailored to the problem of SV we provide a powerful feature extraction mechanism for SV.
2. We apply a standard mechanism for enhancing the verification efficiency of the produced descriptor on meaningful areas of the signature, by means of the key-point detection mechanism of BRISK descriptors.
3. We perform a thorough evaluation of the effects of preprocessing, especially thinning and introduce an automated algorithm to select the optimum thinning level in order to derive a signature's trace on the image plane.
4. We thoroughly investigate the impact of all the major parameters associated with SR, such as selected formulation (greedy approximation or convex relaxation), dictionary size, sparsity level etc. We also evaluate the effect to the overall performance of imposing additional priors into the corresponding optimization problems.

The paper is organized as follows: Section 2 provides a short literature review and summarizes the proposed approach. Section 3 discusses details of the signature preprocessing steps and highlights the feature formation process while Section 4 describes the utilized databases along with the evaluation protocol. Section 5 displays the corresponding experimental results. Finally, conclusions are drawn in Section 6. Appendices A and B provide elementary details regarding sparse representation and dictionary learning techniques while Appendices C and D provide two pseudocodes related to the present work.

## 2 LITERATURE REVIEW - SUMMARY OF THE METHOD

In accordance with Appendices A and B the following notations are introduced: A single patch is a column vector, defined as: $\mathbf{x}^i \in \mathbb{R}^{SignalDim \times 1}$ while matrices are formed by $M$-columns: $\mathbf{X} = [\mathbf{x}^1, \mathbf{x}^{2,}...\mathbf{x}^M] \in \mathbb{R}^{SignalDim \times M}$. The $K$-atoms $\mathbf{d}^j \in \mathbb{R}^{SignalDim \times 1}$ of a dictionary are the columns of $\mathbf{D} = [\mathbf{d}^1, \mathbf{d}^2,...,\mathbf{d}^K] \in \mathbb{R}^{SignalDim \times K}$ matrix with $K > signalDim$ in order to express the overcomplete property. Finally, the matrix $\mathbf{A} = [\mathbf{\alpha}^1, \mathbf{\alpha}^2,...\mathbf{\alpha}^M] \in R^{K \times M}$ represents the sparse coefficients $\mathbf{\alpha}^i \in \mathbb{R}^{K \times 1}$ for a single $i$-patch.

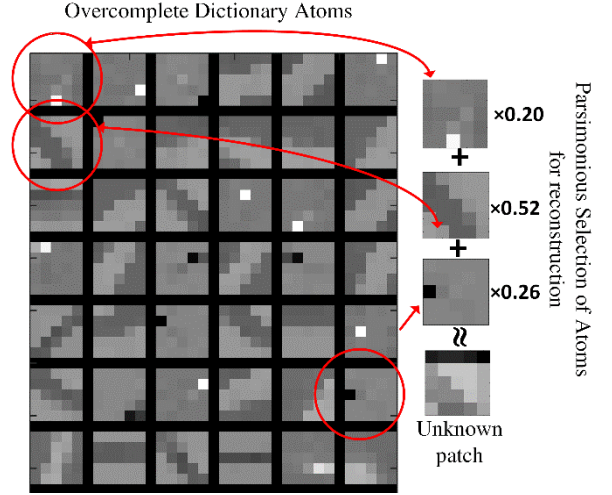

Fig. 1. Illustration of the SR concept. The dictionary is composed of forty-two *atoms* of 25 dimensions (5x5). For an unknown patch, a SR algorithm calculates new representation in the form of the sparsest linear combination of dictionary atoms that minimizes the reconstruction error.

### 2.1. Related Work

A significant number of handcrafted feature extraction methods for offline signature verification rely on the evaluation of global and/or local signature descriptors as well as on grid-texture oriented features. With respect to the above family of feature descriptors, a diversity of feature extraction methods for offline SV has been proposed mainly for single-script [2-6] as well as for multi-script styles of signatures [21, 22]. Some other examples can be found in [19, 20] and their associated references. On the other hand, methods have been proposed that rely on learning features directly from the raw image data. Some efforts include the use of Restricted Boltzmann Machines (RBMs) in [23] and Convolutional Neural Networks (CNNs) [24, 25]. Soleimani et al. in [9] proposed the use of Deep Neural Networks for Multitask Metric Learning by employing a distance metric between pairs of signatures in which Local Binary Patterns is used as an intermediate feature vector. Rantzsch et al. [26] presented an approach named Signature Embedding which is based on deep metric learning. Specifically they compared triplets of two genuine and one forged signature, in order for their system to learn to embed signatures into a high-dimensional space. Following, they proposed a Euclidean distance metric as a means for measuring similarity. Hafemman et al. in a series of publications, proposed methods for learning features from images. Specifically, the authors in [27] introduced a formulation for learning features from genuine signatures by a development dataset, and used them in order to train writer dependent classifiers to another set of users. In [28] the authors obtained state-of-the-art results on several GPDS datasets using CNN architecture and in [29] they demonstrated a novel formulation that leverages knowledge of skilled forgeries for feature learning. In addition, the authors in [8] responded to the fixed size input constraint of the neural network by learning a fixed-sized representation from variable sized signature images with the integration of a spatial pyramid pooling layer.

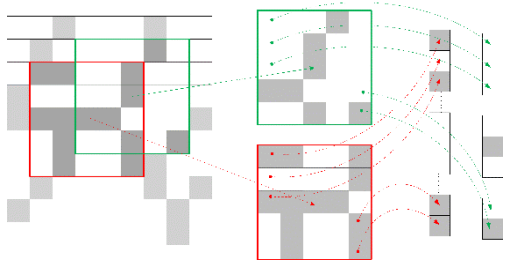

Fig. 2. Patch extraction detail. Left: The gray shades represent pixels of the trace of a signature. Center: Two patch windows of size 5 (5x5 pixels) centered or located at two different points of the signature trace. Right: The formation of the corresponding column-vectors by sequential column-wise concatenation of each patch's values.

Although SR methods have not been exploited for feature extraction in offline SV systems prior to [19, 20], in [30] a method for writer identification based on sparse representation of handwritten structural primitives, called graphemes or fraglets is presented. Similarly, in [31] an online signature verification technique appears based on discrete cosine transform (DCT) and sparse representation. In addition, some sort of codebook formation is also proposed in [32] by means of forming codebooks using signature samples of an independent database with the k-means algorithm and in [13] by creating a codebook of first order HOGs and then coding each feature to the nearest word in the codebook. Complementary, we were also motivated to pursuit sparse coding and dictionary learning instead of k-means, since it has been stated that "whenever using k-means to get a dictionary, if you replace it with sparse coding it'll often work better" [33]. Therefore our intuition to use SR just got stronger.

## 2.2 Method Overview

The proposed handwritten signature verification system utilizes a sparse representation framework in order to learn local features and construct a global signature descriptor. The two main approximations of sparse coding (greedy and convex relaxation) and their efficiency are thoroughly investigated. Various styles of pooling strategies including a novel, second order pooling function are also presented. In addition, a widespread key-point selection algorithm is also employed in order to further emphasize salient locations across the signature.

The preprocessing stage of grayscale signature images is kept as simple as possible given the fact that usually the signatures under examination have already undergone a noise – artefacts removal process. The major steps are: a) thresholding, followed by b) the morphological process of thinning. The Optimal Thinning Level (OTL) of each signature is defined with the use of its local patches, derived from the binary/thinned images. Specifically, patches are extracted from every pixel of the thinned signature's trace (i.e. by dense sampling, as shown in Figure 2) whereas the percentage of the signature pixels that inhabit each patch is also counted. This information helps us to define the patch density (PD) of a signature and plot it as a function of the number of successive thinning operations applied. In this way, the individual optimal thinning level (OTL) for each signature can be defined by utilizing the minimum value of its corresponding PD slope. Subsequently, the optimal thinning level (MOTL) for the reference samples of a writer is defined by their median value. The MOTL value is used for preprocessing of all the signature images that belong to the writer under consideration.

For SR, all the densely sampled patches $\{\mathbf{x}^i\}$ of the signature trace are extracted from the grayscale signature image, at every *i*-position indexed by the skeleton pixels of the signature, obtained via the thinning process. Subsequently, the gray values of the patches associated with the genuine reference samples are transformed into column vectors and used as input $\mathbf{X} = \{\mathbf{x}^i\}$ to a dictionary learning algorithm which evaluates the dictionary $\mathbf{D}$. Following, for every other signature, the patches are encoded, by means of evaluating the sparse coefficients $\mathbf{A} = \{\mathbf{a}^i\}$ using the dictionary D of the writer under consideration. The final feature of each signature image is formed by applying a pooling function F($\mathbf{A}$) on the sparse vectors, using a specially designed spatial pyramid, which segments the signature skeleton in equimass parts. Additionally, the key-points derived from the BRISK [34] algorithm pinpoint image regions of interest, whereas the sparse codes of the corresponding nearby patches are also pooled together. The obtained pooled vector is concatenated with the spatial pyramid vector in order to form the final signature descriptor. The signature verification system is realized by a binary radial basis SVM classifier with the above features as inputs. The learning stage (i.e. training and validation) of the verifier, utilizes the positive class features derived from the genuine reference signatures as well as the negative class features derived from randomly selected genuine signatures from a subset of the remaining writers of the dataset. Figure 3 presents a motif of the proposed system with emphasis on the preprocessing and the dictionary learning stages. The corresponding pseudocode is provided in Appendix C.

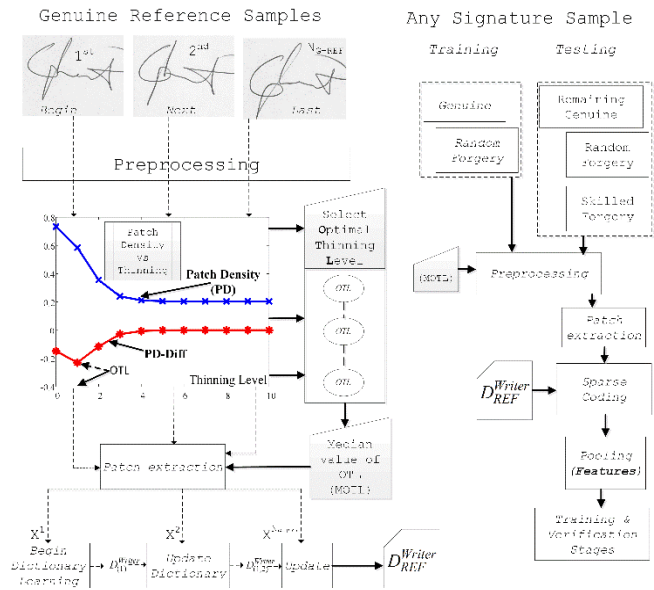

Fig. 3. A diagram of the proposed pipeline emphasizing the preprocessing and dictionary learning stages. The Optimal Thinning Level (OTL) at the preprocessing stage is selected from the extreme point of the patch density slope (located at the upper center of the Figure).

## 3. FEATURE EXTRACTION

### 3.1. Preprocessing

Preprocessing consists of two steps: binarization followed by thinning. The grayscale images are binarized using Otsu's method [35]. Then, successive morphological thinning operations are applied on the binary image in order to provide a gradual skeletonization of the signature. Intuitively, the outcome of the thinning operation affects the verification performance since it modifies the shape of the signature image. It has been experimentally observed [7] that the thinning level is critical for a Signature Verification (SV) system's performance and its optimal value is not the same for all databases. This work proposes a novel method for selecting the optimal thinning level (OTL) for each signature and consequently for each writer. As mentioned in Section 2.2, the OTL is defined as the number of thinning operations that results to the steepest descend of the density function. Following the enrollment of a set of genuine reference signatures for a person, its median value of the associated OTL values (MOTL) accompanies the design of each writer's model, i.e. $MOTL(N_{G-REF}) = median(OTL(i))$, where $i \in [1,...,N_{G-REF}]$. Hence, for any input signature which claims an identity, the number of thinning operations will be determined by the MOTL value of the signing person. Appendix D presents the proposed thinning procedure in an algorithmic form. Figure 4 presents the corresponding plots of the patch density, the patch density derivative, expressed by its associated patch density difference and OTL-MOTL values for one writer and an indicative number of his/hers genuine samples derived from all the signature datasets, namely CEDAR, MCYT-75, GPDS300 and UTSIG.

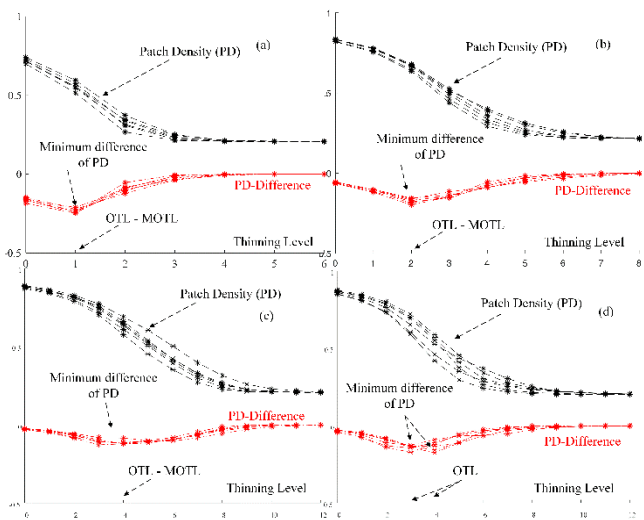

Fig. 4. Plots of the patch density (PD) and corresponding differences as a function of the thinning level. The above plots refer to one writer and an indicative set of six genuine samples derived from the (a) CEDAR, (b) MCYT-75, (c) GPDS300 and (d) UTSIG signature datasets, respectively. The patch size is set to 5.

The extracted results indicate that the genuine signatures that are part of the CEDAR dataset have the majority (~95%) of their OTL values to be equal to one with few (~5%) signature samples having their OTL values to two. For the MCYT-75 dataset the majority of their OTL values are equal to two with very few samples having their OTL values to three and four. For the GPDS300 dataset, the OTL values appear to distribute between 2 and 6, with an observed mode of 4, while for the UTSIG dataset OTL is equally distributed among values 2 and 3. Especially for the CEDAR and MCYT-75 datasets, Figure 4 indicates that they are more stable in terms of acquisition conditions comparing to datasets GPDS300 and UTSIG. Thus one could consider the SV problem addressed by CEDAR and MCYT-75 dataset to resemble a case study in which the aspect ratio and acquisition conditions do not vary significantly, similar to situations encountered in mobile financial applications.

### 3.2. Patch Extraction

The signature patches are extracted from the original grayscale signature image, indexed by the signature's skeleton pixels after applying the thinning operation MOTL times, as shown in Figure 4. Specifically, the patches' centers are sampled densely at every pixel of a signature's skeleton. As a consequence, the number of image patches equals the number of pixels of the signature's skeleton. Furthermore, the patches are centered, i.e. have their average intensity been subtracted in order to have a zero mean value. The centering of each patch produces data invariant to the mean intensity and the learned structures, like edges, are anticipated to have zero mean as well. In all the conducted experiments the patch size is set to five; thus the mask-patch has a dimensional size of 25 ($n=5\times5$). Our main rationale behind this selection is to keep the complexity of the local manifold of patches reasonably low. This is because sparse codes of data lying on smoother and more uniform manifolds tend to create more evenly distributed coefficients along the codebook's elements. Such cases are less prone to the phenomenon where very few dictionary elements are over-represented in the resulting sparse codes and are responsible for a significant amount of the total energy. Thus, they predominantly shape the distance between global image descriptors formed by these codes and require special pooling strategies to restore the discriminative power [36].

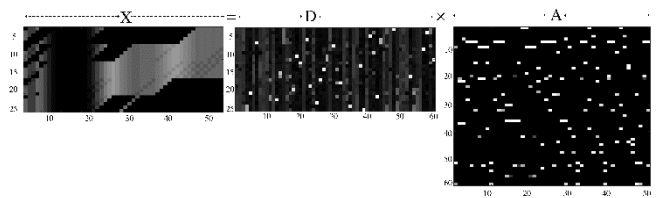

Fig. 5. Graphical illustration for signature coding with the use of SR. In this example, the matrix **X** consists of fifty 25-dimensional column-wise patches $\mathbf{x}^i$ of associated signature pixels. The lexicon **D**, consists of sixty atoms and is previously learned via a dictionary learning algorithm. The patch matrix **X** is represented as a linear combination of elements from **D**, with sparse coefficients given by **A**. Notice the sparse nature of **A** with few lines activated (appeared as

TABLE 1
OPERATIONAL PARAMETERS FOR SR METHODS

| Process | Parameter | symbol | Value |
|---|---|---|---|
| Thinning | patch size | - | 5 |
| K-SVD / OMP | # maximum iterations | $t_{max}$ | 50 |
| | # atoms | K | 60 |
| | Sparsity level | $\rho$ | 3 |
| Online / LARS-Lasso | mini-batch size | $\eta$ | 512 |
| | # atoms | K | 60 |
| | regularization parameter | $\lambda$ | 0.15 |

With this aim, it is valuable to consider the parameters which affect the dimensionality and shape of the underlying local manifolds. In [37] Peyré shows that the local manifold of patches from cartoon images (images that contain sharp variations along regular curves) can be parameterized by two variables, leading to a manifold topologically equivalent to the surface of a cylinder in 3D space. This parameterization holds as long as the signal within each patch can be approximated by two regions (black and white) separated by a linear segment. If the patch size become larger and the edges within the patches appear curved, extra degrees of freedom have to be included to the signal's model thus leading to a more complex manifold. Similarly to cartoon images, the nature of the signal within signature patches is such that can be modeled by a handful of parameters if the patch size is small-enough, indicating a low-dimensional underlying manifold structure. On the other hand, the complexity can be dramatically increased if the patch size becomes large enough to contain curves and parts from neighboring line segments. We set patch size equal to 5, since it is a good tradeoff between the underlying signal's complexity since for smaller patches the local manifold obviously becomes degenerate - and the overall computational complexity. The later is dictated by the dictionary size whose over-completeness requirement points back to the patch dimensionality as the most significant parameter. It is worth noting that this scale of patches has also proven to be efficient in previous research efforts [7] on the particular problem, delivering state-of-the-art results and strengthening our decision on this selection.

### 3.3. SR-driven local feature extraction

Sparse Coding (SR) and consequent Dictionary Learning is a popular technique for handling computer vision problems. In this work, SR is involved in order to construct the dictionary (or **D** for simplicity) that represents the characteristic properties of the signatory. The patches extracted from one signature are represented as columns of the matrix $\mathbf{X} \in R^{n \times M}$ and the dictionary for sparse representation is updated via an iterative process, in which the patches of one reference signature are used in each update and ultimately, all the reference signatures are utilized in a cascade fashion. Hence, dictionary is updated consecutively using each one of the writer's genuine reference signatures. Thus, the system can integrate easily a new reference genuine signature of the writer and thus, it is practical for application in everyday scenarios. Following the construction of the dictionary, for any other inserted signature image its patches are extracted and encoded using the dictionary and SR coding in order to obtain the sparse representation matrix **A**. Figure 5 presents the SR concept as it is applied in the proposed method for signature modeling. Depending on the case, the dictionary learning and the sparse coding stages are implemented with either the K-SVD/OMP or the SPAMS/LARS-Lasso algorithm pairs. In addition, the impact on the system performance of some other popular optimization constraints is also explored with priors such as the positivity constraint of the coefficients, the non-negativity constraint for the dictionary atoms and the non-negative matrix factorization (NMF) method in which the matrix **D** and the vectors **A** are required to be positive. For the cases in which **D** is positive, the SPAMS/LARS-Lasso algorithm is invoked but without the centering procedure of patches. It must be stated here that the centering procedure has not been applied for the case of having positive dictionary atoms and the NMF.

The operating parameters of the KSVD/OMP and the online/LARS-Lasso algorithms are summarized in Table 1. Specifically, for the KSVD/OMP case, the number of maximum iterations $t_{max}$ is set to fifty, as this number has been found experimentally that is adequate for all cases. The number of atoms $K$ was set to sixty in order to ensure the over-completeness of the dictionary, which follows the rule of thumb that suggests a number of atoms more than twice the dimension of the patch size ([38], pg. 33). The sparsity level $\rho$ is set to three in order to provide an overall of 5% sparsity on the coefficients given the fact that when $\rho$ is small enough relative to $K$, the approximating methods are known to perform very well [39]. For the online/LARS-Lasso case, we use the same number of atoms $K$ as in the KSVD/OMP case, while the number of signals drawn at each iteration or mini-batch size $\eta$, is set to 512, in order to improve the efficiency of the online dictionary learning algorithm ([38], pg.168). The dictionary learning algorithm is allowed to run for a typical execution time of one minute, which is considered adequate given the size of the signature patches. In addition, the regularization parameter $\lambda$ is set to 0.15, a value which is in proximity to the classical $1/\sqrt{n}$ normalization factor proposed by Bickel et al. [40].

These parameters are a-priori fixed for all datasets and signatories and are not defined or tuned with neither a validation or test set. We should notice that we do not tune these parameters with a test set. In addition, we experimentally found that it is almost useless to consider a validation set, consisted of genuine and random forgeries, in order to optimize further any of the aforementioned parameters. That would make sense only in case where skilled forgeries are employed; however this protocol, in the WD training context, is usually not followed.

### 3.4. Pooling Strategies

It is common for contemporary computer vision algorithms to incorporate a pooling stage, which aggregates local features over a region of interest [41-43]. In this work, a number of variants are proposed as signature descriptors by global aggregation of the local patch sparse coding coefficients into a final vector, through an appropriate pooling function. Depending on the pooling function, the corresponding signature descriptor is denoted hereafter as $f_\mathbf{I}^{F1}$ - $f_\mathbf{I}^{F5}$, and defined as follows:

(F$_1$): $f_\mathbf{I}^{F1} = \{f_\mathbf{I}^{F1}[j]\} = \{\frac{1}{M}\sum_{i=1}^{M}\boldsymbol{\alpha}^i[j]\}, j=1:K$ (1)

(F$_2$): $f_\mathbf{I}^{F2} = \{f_\mathbf{I}^{F2}[j]\} = \max(|\boldsymbol{\alpha}^i[j]|), i=1:M, j=1:K$ (2)

(F$_3$): $f_\mathbf{I}^{F3} = \{f_\mathbf{I}^{F4}[j]\} = \{\sqrt{\frac{\sum_{i=1}^{M}(\boldsymbol{\alpha}^i[j]-f_\mathbf{I}^{F1}[j])^2}{M-1}}\}, j=1:K$ (3)

(F$_4$): $f_\mathbf{I}^{F4} = \{f_\mathbf{I}^{F4}[j]\} = \frac{\sum_{i=1}^{M}\boldsymbol{\alpha}^i[j]}{\sum_{j=1:}^{K}\sum_{i=1}^{M}\boldsymbol{\alpha}^i[j]}, j=1:K$ (4)

(F$_5$): $f_\mathbf{I}^{F5} = \{f_\mathbf{I}^{F5}[j]\} = \frac{\sum_{i=1}^{M}\boldsymbol{\alpha}^i[j]}{\sqrt{\sum_{j=1:}^{K}(\sum_{i=1}^{M}\boldsymbol{\alpha}^i[j])^2}}, j=1:K$ (5)

Average pooling (F$_1$) is the simplest pooling function that estimates the average SR coefficients from the whole region of interest. The max pooling (F$_2$) operation only captures the most salient representation value from the entire region of interest. Standard deviation (F$_3$) is proposed here as an alternative pooling function that captures 2$^{nd}$ order statistics of the coefficients' distribution, in an aim to investigate if this type of information can deliver better discrimination capabilities to the resulting descriptor. Normalized Sum pooling (F$_4$) function produces vectors with intensity invariance, and finally the (F$_5$) function produces $l^2$ normalized vectors projected onto the unitary ball, which can be important for linear classification kernels [41].

The final feature vector's dimensionality provided by the preceding pooling operations is a multiple of the number of dictionary atoms K. In order to encapsulate local signature information to the final feature vector, a specially designed spatial pyramid is employed for segmenting the signature images into a grid of $(\beta \times \beta)$ equimass sub-regions. For each $I_\beta(t), t=1:\beta^2$ segment of this pyramid, its selected $\mathbf{x}_{I_{\beta^2}(t)}$ patches are enabled for indexing the local $\boldsymbol{\alpha}_{I_{\beta^2}(t)}$ representation coefficients, which in their turn are subjected to the same pooling operations $F(\mathbf{A}_{I_{\beta^2}(t)})$ that is used for the computation of the $f_\mathbf{I}^{F1}$ - $f_\mathbf{I}^{F5}$ global versions. As an aftermath, the dimensionality of the expanded feature $f^{F_g} = \{f_\mathbf{I}^{F_g}, f_{I_{\beta^2}}^{F_g}\}, g=1:5$, now equals to $(\beta^2+1) \times K$. In this work we tested the system's performance for two values of the $\beta$ parameters, i.e. $2 \times 2$ and $3 \times 3$ equimass sub-regions for the spatial pyramid.

### 3.5. Emphasizing on informative local keypoints

A human expert, who wishes to analyze a signature image in order to verify if it is genuine or forgery, focuses on certain points of the signature. In an effort to discover these signature's points of interest we utilize the saliency-modeling mechanism implemented by the Binary Robust Invariant Scalable Keypoints (BRISK) [34]. BRISK have the advantage of dramatically lowering computational complexity and thus are suited for low power devices, such as practical portable SV systems. BRISK computation relies on an easily configurable circular sampling pattern from which it computes brightness comparisons to form a binary 512 bit descriptor string and estimates keypoint scale in continuous scale-space. In this work only the detected keypoint locations and not the keypoint descriptors (BRISK) are utilized. Hence, keypoints indicate the patches, and thereafter the corresponding sparse codes will be once more pooled together in order to obtain an additional feature vector. This vector is concatenated with the spatial pyramid vector resulting to a final feature vector $f^{F_g} = \{f_\mathbf{I}^{F_g}, f_{I_{\beta^2}}^{F_g}, f_{I_{BR}}^{F_g}\}$ of dimensionality is $(\beta^2+2) \times K$. Figure 6 depicts a zoomed area of an example signature image where the BRISK keypoints and their nearest signature pixels are denoted as crosses and x-points, respectively.

We provide a visualization, of the way that the signatures are distributed in the feature space by employing the t-SNE algorithm [44], [29] in order to map the genuine references, remaining genuine, skilled forgeries and random forgery samples from the initial R$^{300}$ dimensional space to the projected R$^2$ space. Figure 7, displays the aforementioned information for F$_3$ type of feature, for one user and for the GPDS dataset.

## 4. EVALUATION

The selected classifier is a binary radial basis SVM classifier. In the learning stage of the classifier, each one of the $N_{G-REF}$ genuine reference samples is encoded with the OMP or the LARS-Lasso SR algorithms in order to provide the positive class $\omega^\oplus \in R^{N_{G-REF} \times (Dim)}$ for each one of the feature $f_\mathbf{I}^{F1}$ - $f_\mathbf{I}^{F5}$ descriptor given in (1)-(5). The negative training class $\omega^- \in R^{2 \cdot N_{G-REF} \times (Dim)}$ is composed by taking $2 \times N_{G-REF}$ random forgery samples; that is one genuine sample from a population of $2 \times N_{G-REF}$ writers different than the examined one. The learning feature set population $[\omega^\oplus ; \omega^-]$ is partitioned during the training / cross-validation procedure into a corresponding training and holdout validation set which serves as an input to the classifier. The holdout cross-validation procedure returns the optimal values of the $C^{opt}$ and gamma-$\gamma^{opt}$ parameters with respect to a maximum cross validation value of the associated Area Under Curve [45]. In addition, the cross-validation procedure stores the corresponding output scores $CVS^\oplus$ which are conditioned upon the positive $\omega^\oplus$ class samples. The testing stage utilizes the remaining genuine samples, the skilled forgeries along with a number of random forgeries.

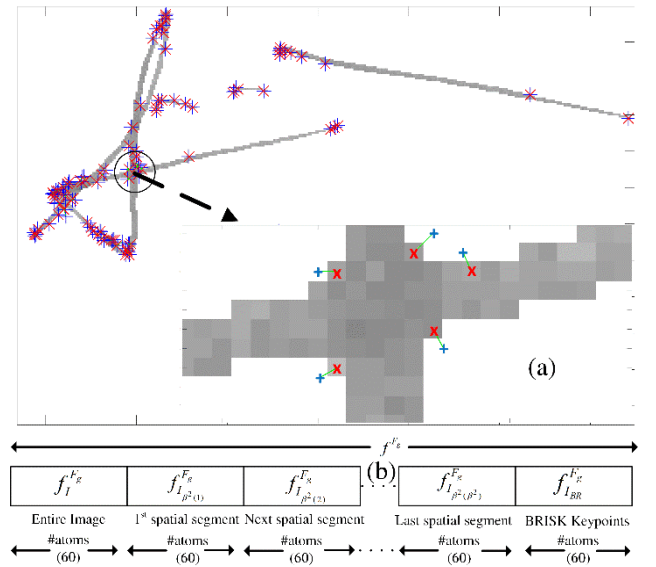

Fig. 6. a) Detail of a signature with BRISK keypoints. In the detail image, the marker (+) indicates the BRISK actual coordinates while (x) mark the assigned nearest signature pixel neighbors. b) Configuration of the various $f^{F_g}$ in the overall feature vector. In this example, the number of atoms has been set to 60.

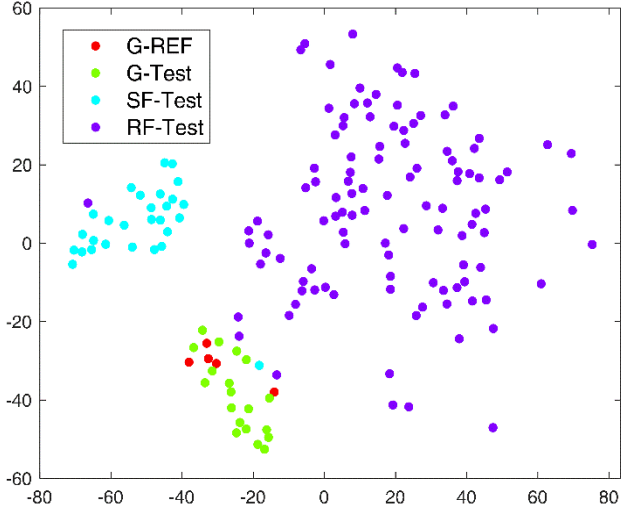

Fig. 7. t-SNE projection of the embedding space for the genuine reference (G-REF), remaining genuine (G-Test), skilled forgeries (Sf-Test) and some random forgery samples (RF-test) for $F_3$ feature type.

### 4.1. Datasets and Experimental Protocols

Four signature datasets are used in order to test the proposed system architecture. The first one is the popular CEDAR dataset [46]. For each one of the 55 enrolled writers, a total of 48 signature specimens (24 genuine and 24 simulated) confined in a 50 mm by 50 mm square box are provided and digitized at 300 dpi. The simulated signatures found in the CEDAR dataset are composed from a mixture of random, simple and skilled forgeries. The second signature dataset is the off-line version of the popular MCYT signature database [47, 48]. A whole of 30 signature samples (15 genuine and 15 simulated) signature samples are recorded for each one of the 75 enrolled writers at a resolution of 600 dpi and for and the capture area is $127 \times 97$mm. Both CEDAR and MCYT-75 datasets have their samples confined within one bounding box. The third signature dataset is the GPDS300 [48, 49] which contains 24 genuine signatures and 30 simulated forgeries of 300 individuals stored in an 8-bit, grey level format. A special feature of this dataset is that the acquisition of signature specimens is carried out with the aid of two different bounding boxes of size $5 \times 1.8$ cm and $4.5 \times 2.5$ cm respectively. As a result, the files of this dataset include images having two different aspect ratios; this phenomenon conveys a structural distortion highlighted during the feature extraction procedure. The fourth signature dataset is the Persian UTSIG, created by Soleimani et al. [50]. It contains specimens from 115 writers where each one has 27 genuine signatures, 3 opposite-hand signatures, and 42 skilled forgeries made by 6 forgers. As stated by the dataset creators, a property of the UTSIG dataset, compared with the other public and popular ones, is that UTSIG has more samples, more classes, and more forgers. An important characteristic of the UTSIG is that the acquisition of signature specimens is carried out with the aid of six different bounding box sizes simulating real world conditions and public services application forms. Since this work addresses writer depended signature verification, for any signatory a specific model is created by randomly employing $N_{G-REF}$ genuine reference signature samples. The number of $N_{G-REF}$ is primarily set to 5 for addressing cases in which only a few samples are available. In order to provide comparable results with other state-of-the-art methods, we allowed the $N_{G-REF}$ parameter to assume values from 5 and/or 12 according to the specific needs of each dataset. The remaining genuine samples of a writer, as well as its forgeries and other writers' genuine samples have been used in order to form the test set.

A number of methods exist in order to quantify the efficiency of the proposed system. The false acceptance rate FAR and its associate probability $P_{FAR}$ depicts a system's measure of the resistance to input samples other that the genuine ones like random or skilled forgeries. On the other hand, the FRR - false rejection rate and its accompanied probability rate $P_{FRR}$ provides a system's measure of the failure to genuine samples. These operating system parameters are computed as a function of a sliding threshold whose extremes are located between the minimum and maximum values of the cross-validation SVM output scores $CVS^\oplus$. Some examples include independent experiments and corresponding solutions like: a) the $P_{FAR}^S$ vs. $P_{FRR}$, where the upper script S denotes skilled forgery, along with their average $AVE^S$ or equal error rates:

TABLE 2
VERIFICATION ERROR RATES (%) FOR THE CEDAR AND MCYT-75 SIGNATURE DATASETS WITH $L_0$-NORM: KSVD/OMP DICTIONARY/SR METHODS. NUMBER OF REFERENCE TRAINING SAMPLES EQUALS 5. THE $EER^S$ RATE CORRESPONDS TO THE USER SPECIFIC THRESHOLD CASE.

| SET | F(A) | SP: $\beta=2$ (Dim=360) | | | | SP: $\beta=3$ (Dim=660) | | | |
|---|---|---|---|---|---|---|---|---|---|
| | | $P_{FAR}^S$ | $P_{FRR}$ | $EER^S$ | $P_{FAR}^{R@EER^S}$ | $P_{FAR}^S$ | $P_{FRR}$ | $EER^S$ | $P_{FAR}^{R@EER^S}$ |
| CEDAR | F1 | 6.83 | 7.32 | **2.67** | 0.43 | 6.13 | 7.18 | **2.80** | 0.27 |
| | F2 | 9.45 | 9.83 | **4.17** | 2.93 | 8.83 | 8.65 | **3.21** | 0.76 |
| | F3 | 4.76 | 4.91 | **1.44** | 0.17 | 4.78 | 5.63 | **1.80** | 0.12 |
| | F4 | 8.12 | 9.94 | **7.51** | 1.95 | 12.7 | 20.2 | **13.4** | 4.82 |
| | F5 | 5.31 | 5.27 | **3.08** | 0.35 | 5.05 | 5.11 | **2.48** | 0.24 |
| MCYT-75 | F1 | 10.4 | 7.32 | **3.80** | 0.22 | 9.18 | 7.12 | **3.19** | 0.14 |
| | F2 | 14.7 | 14.6 | **10.9** | 3.16 | 15.6 | 11.5 | **8.65** | 1.12 |
| | F3 | 8.09 | 6.75 | **3.18** | 0.16 | 8.07 | 6.21 | **2.82** | 0.07 |
| | F4 | 12.99 | 16.17 | **10.8** | 3.08 | 11.0 | 23.7 | **16.7** | 6.17 |
| | F5 | 7.64 | 7.76 | **3.23** | 0.14 | 7.77 | 7.92 | **3.46** | 0.13 |

## TABLE 3
VERIFICATION ERROR RATES (%) FOR THE CEDAR AND MCYT-75 DATASETS WITH $L_1$-NORM: PARAMETERS AS IN TABLE 1.

| SET | F(A) | SP: $\beta = 2$ (Dim=360) | | | | SP: $\beta = 3$ (Dim=660) | | | |
|---|---|---|---|---|---|---|---|---|---|
| | | $P_{FAR}^S$ | $P_{FRR}$ | $EER^S$ | $P_{FAR}^{R@EER^S}$ | $P_{FAR}^S$ | $P_{FRR}$ | $EER^S$ | $P_{FAR}^{R@EER^S}$ |
| CEDAR | F1 | 6.99 | 7.60 | **2.65** | 0.37 | 6.59 | 6.61 | **3.08** | 0.23 |
| | F2 | 9.42 | 9.71 | **3.45** | 1.55 | 8.52 | 8.01 | **2.80** | 0.48 |
| | **F3** | 4.61 | 4.64 | **1.62** | **0.12** | 4.88 | 5.81 | **2.01** | **0.08** |
| | F4 | 8.19 | 8.73 | **3.52** | 0.56 | 9.17 | 8.71 | **3.97** | 0.41 |
| | F5 | 7.05 | 7.81 | **3.38** | 0.41 | 5.49 | 6.48 | **3.07** | 0.22 |
| MCYT-75 | F1 | 10.46 | 8.57 | **3.91** | 0.24 | 7.82 | 7.68 | **3.58** | 0.14 |
| | F2 | 15.8 | 14.3 | **11.20** | 3.09 | 13.9 | 12.3 | **8.60** | 1.08 |
| | **F3** | 8.35 | 7.58 | **3.71** | **0.22** | 7.59 | 7.29 | **3.40** | **0.10** |
| | F4 | 12.2 | 13.43 | **3.93** | 0.30 | 9.82 | 9.68 | **3.53** | 0.15 |
| | F5 | 7.87 | 7.59 | **3.66** | 0.20 | 7.07 | 7.69 | **3.05** | 0.12 |

$EER^S : P_{FAR}^S = P_{FRR}$, and b) the $P_{FAR}^R$ vs. $P_{FRR}$, where the upper script R denotes random forgery, along with their corresponding average $AVE^R$ or equal error rates $EER^R : P_{FAR}^R = P_{FRR}$. Other proposals provide joint solutions [51], which initially evaluate the $TH^{@EER^S}$ - threshold, defined as the value which designates the $EER^S$ operating point and then, taking into account this value in order to assess the $P_{FAR}^{R@EER^S}$ error rate.

Regarding the choice of metrics, several researchers suggest handling SV as either a one class pattern recognition problem or a two class pattern recognition problem. The issue arises from the fact that the negative class of the test set has representatives from both skilled and random forgery samples [52]. The key point is that the skilled forgery class is composed of few samples compared with the random class. So, if one tries to incorporate both forgery populations there is always the danger of reporting biased results. For example, reporting an EER point of $EER^{R\&S} : \{P_{FAR}^{R\&S}\} = P_{FRR}$ where all types of forgery samples are employed; this is clearly biased since the forgery class has samples with two different populations of forgery. In this paper, as in [19, 20] the $EER^S_{user\text{-}threshold}$ is used by employing user-specific decision thresholds in order to evaluate the verification performance of the proposed system.

## TABLE 5
VERIFICATION ERROR RATES (%) FOR THE GPDS300 AND UTSIG SIGNATURE DATASET WITH $L_0$-NORM: NUMBER OF TRAINING SAMPLES EQUALS 5 AND [5,12] FOR THE GPDS AND UTSIG.

| SET | F(A) | SP: $\beta = 2$ (Dim=360) | | | | SP: $\beta = 3$ (Dim=660) | | | |
|---|---|---|---|---|---|---|---|---|---|
| | | $P_{FAR}^S$ | $P_{FRR}$ | $EER^S$ | $P_{FAR}^{R@EER^S}$ | $P_{FAR}^S$ | $P_{FRR}$ | $EER^S$ | $P_{FAR}^{R@EER^S}$ |
| GPDS300 | F1 | 6.91 | 6.22 | **2.47** | 0.59 | 6.46 | 6.47 | **3.14** | 0.44 |
| | F2 | 8.27 | 8.89 | **3.98** | 3.70 | 11.79 | 9.37 | **4.43** | 1.82 |
| | **F3** | 6.73 | 6.13 | **1.50** | 0.33 | 5.01 | 5.92 | **1.97** | 0.19 |
| | F4 | 16.1 | 19.3 | **15.2** | 6.23 | 18.6 | 29.4 | **20.4** | 10.0 |
| | F5 | 7.79 | 7.96 | **3.30** | 0.55 | 7.58 | 6.83 | **3.53** | 0.30 |
| UTSIG-5G | F1 | 18.3 | 16.9 | **11.7** | 1.91 | 16.9 | 12.4 | **9.82** | 1.94 |
| | F2 | 25.3 | 21.6 | **17.0** | 7.02 | 23.6 | 15.7 | **13.35** | 3.33 |
| | **F3** | 16.1 | 13.4 | **9.94** | 1.27 | 15.8 | 12.5 | **8.56** | 1.25 |
| | F4 | 24.5 | 27.6 | **21.1** | 6.39 | 29.8 | 29.5 | **25.15** | 12.5 |
| | F5 | 20.7 | 24.6 | **12.2** | 1.48 | 17.3 | 15.2 | **10.48** | 0.92 |
| USTIS-12G | F1 | 13.6 | 13.5 | **8.13** | 0.53 | 12.1 | 9.37 | **7.33** | 0.25 |
| | F2 | 19.9 | 18.1 | **12.0** | 3.66 | 18.9 | 12.3 | **11.4** | 0.87 |
| | **F3** | 13.2 | 11.5 | **7.36** | 0.33 | 9.77 | 8.17 | **6.22** | 0.12 |
| | F4 | 23.8 | 24.6 | **17.3** | 3.32 | 29.3 | 21.1 | **20.1** | 6.63 |
| | F5 | 13.9 | 12.7 | **8.51** | 0.39 | 10.1 | 9.55 | **7.44** | 0.21 |

Also the calculation of the $P_{FAR}^S$, $P_{FRR}$ rates with the utilization of a predetermined threshold (i.e. hard decisions) is provided by using the a-priori knowledge of the cross validation procedure scores $CVS^{\oplus}$. Specifically, the hard threshold value corresponds to the 50% of the average of the genuine $CVS^{\oplus}$ scores for each writer. For completeness, at this specific threshold point the $P_{FAR}^R$ error rate is evaluated by employing the genuine samples of the remaining writers from the test set.

## 5. EXPERIMENTAL RESULTS

For the sake of sanity and in order to avoid exhausting tests on all datasets, complete results are provided in tables 2-3 which involve the popular CEDAR and MCYT-75 datasets. All the experimental protocols are repeated ten times and the results are averaged in order to provide meaningful comparisons. It can be noticed that the verification error rates for the CEDAR and MCYT datasets do not vary substantially when both $l_0$ and $l_1$-norms dictionary learning and SR are used. This is in accordance to the material exposed in Appendices A and B as well as in Section 3.3 which states that for a given dictionary **D** both $l_0$ and $l_1$ oriented SR solutions are equivalent, for some specific values of the design parameters $\rho$ and $\lambda$. Of course, the problem of locating the appropriate value for these design parameters is not a trivial one since it depends on the individual signature characteristics. An additional issue that arises is that tuning the system to a specific value either for $\rho$ or $\lambda$ depends mainly upon the different training modes. Moreover, we observed during the conducted experiments that the cross validation procedure, which is used for the selection of the optimal classifier parameters, is almost ineffective against skilled forgeries. A potential solution could emerge by employing WI systems with skilled forgeries which are not assigned to any specific person during the training. However, this approach is out of the scope of the present work.

The experimental outcomes show that the best results for the CEDAR dataset are obtained with the use of a spatial pyramid with $\beta = 2$ while for the MCYT-75 the optimal results are obtained with the use of a spatial pyramid with $\beta = 3$. This is probably due to the fact that CEDAR signature specimens have been scanned with a resolution of 300dpi, while the MCYT-75 ones with a resolution of 600dpi, thus it is natural to expect that more pixels exist in the MCYT-75 segments. It should be noticed that, for the CEDAR dataset even when the spatial pyramid with $\beta = 2$ is used, there are a few patches in some segments, which profoundly did not provide any sort of discriminatory information, something that has not been encountered in the case of the MCYT-75.

An important result that arises from our experimental results is that the accuracy for the $F_3$ pooling function seems to outperform all other pooling functions in almost all cases. It is also interesting that although the $F_2$ feature performs quite well in other computer vision applications, this is not the case for signatures images. A possible explanation relies on the very nature of the signature images which are a particular class of image signals that exhibit a degenerate structure. Since all signature images essentially share a limited set of structural elements, the resulting sparse coefficients of different signatures may not differ sufficiently in a 1st order statistic sense, as being the case in more complex image structures. Therefore, it is reasonable to deduce that higher-order statistics can deliver a better level of discrimination between the respective distributions of sparse coefficients, resulting into better verification performance.

TABLE 4
Verification error rates (%) for the CEDAR and MCYT-75 signature dataset with $L_1$-norm: and for the following priors a) positivity constraint of the **A** coefficients, b) the 'non-negative' **C'** constraint for the dictionary atoms and c) the NMF method. $\beta_{CEDAR} = 2$, $\beta_{MCYT} = 3$.

| SET | F(A) | A positive | | | | C' dictionary constrains | | | | NMF | | | |
|---|---|---|---|---|---|---|---|---|---|---|---|---|---|
| | | $P_{FAR}^S$ | $P_{FRR}$ | EER$^S$ | $P_{FAR}^{R@EER^S}$ | $P_{FAR}^S$ | $P_{FRR}$ | EER$^S$ | $P_{FAR}^{R@EER^S}$ | $P_{FAR}^S$ | $P_{FRR}$ | EER$^S$ | $P_{FAR}^{R@EER^S}$ |
| CEDAR | F1 | 7.26 | 7.99 | **2.38** | 0.29 | 7.29 | 7.72 | **2.43** | 0.34 | 9.13 | 8.41 | **2.22** | 0.39 |
| | F2 | 9.56 | 9.07 | **3.19** | 1.52 | 9.32 | 9.84 | **3.63** | 1.72 | 22.5 | 24.2 | **16.3** | 13.3 |
| | F3 | 4.74 | 4.92 | **1.77** | 0.15 | 4.97 | 4.96 | **1.71** | 0.26 | 6.48 | 9.41 | **1.52** | 1.09 |
| | F4 | 10.3 | 9.69 | **3.91** | 0.45 | 10.3 | 9.94 | **3.00** | 0.45 | 12.7 | 10.3 | **3.05** | 1.04 |
| | F5 | 9.98 | 7.69 | **3.17** | 0.35 | 9.83 | 9.99 | **2.78** | 0.32 | 9.78 | 9.36 | **2.64** | 0.54 |
| MCYT-75 | F1 | 12.6 | 8.96 | **4.71** | 0.31 | 11.1 | 10.6 | **5.53** | 0.44 | 10.2 | 10.9 | **4.34** | 0.45 |
| | F2 | 19.4 | 18.3 | **12.4** | 3.43 | 18.8 | 19.4 | **11.9** | 2.50 | 30.8 | 30.4 | **24.3** | 20.3 |
| | F3 | 12.2 | 8.44 | **4.97** | 0.22 | 10.9 | 9.67 | **4.86** | 0.33 | 13.4 | 12.9 | **7.99** | 1.78 |
| | F4 | 14.4 | 11.3 | **5.38** | 0.59 | 13.6 | 14.1 | **5.34** | 0.45 | 17.2 | 18.1 | **6.69** | 1.98 |
| | F5 | 11.3 | 10.2 | **4.69** | 0.35 | 11.4 | 10.2 | **4.97** | 0.27 | 11.8 | 1.6 | **4.48** | 0.37 |

For further investigation we selected the spatial segmentation that corresponds to the best results, i.e. $\beta = 2$ for the CEDAR and $\beta = 3$ for the MCYT-75, and the $l_1$-norm. Table 4 provides comparisons with exactly the same learning and testing samples for the following cases: a) the positivity constraint of the **A** coefficients, b) the 'non-negative' $C'$ constraint for the dictionary atoms and c) the NMF method. The results indicate that none of these cases seems to provide any sort of significant improvement to the verification error. Thus, for the GPDS300 signature dataset and given the $l_1/l_0$ equivalence we will remain to the ($l_0$) norm expressed by the KSVD/OMP algorithms. Table 5, provides the corresponding results. As it is observed, F$_3$ still outperforms all other pooling functions. The increase of the spatial pyramid size from two to three does not provide any considerable verification improvement, except for the random forgery error as it is expressed from the $P_{FAR}^R$ rate. We believe that this effect can be explained due to the indiscriminately nature of the spatial pyramid equimass segmentation which does not take into account the two different aspect ratios of the bounding box. As mentioned earlier, the UTSIG dataset is, according to the author's opinion, a significant realization step towards the assessment of situations which resemble typical conditions and constraints that are broadly encountered in daily transactions. Table 5 also presents the results for our usual case of $N_{G-REF}$ =5. As expected, the provided verification results are poor regarding to the rates of the previous datasets however a closer look to $P_{FAR}^R$ indicates that most likely the $N_{G-REF}$ =5 is not adequate in this case due to the large number of bounding box sizes. Therefore, in accordance with the literature [39], [9] and for comparison purposes, we let the value of $N_{G-REF}$ raise up to twelve. Again, Table 5 presents the corresponding results. Table 5 provides evidence that the evaluation metrics rate drops significantly when the number of reference samples increases. Following and for the sake of simplicity, Table 6 shows comparative results by means of the $EER^S_{user-threshold}$ only for the case in which each input signature is thinned not by the MOTL value of the claimed writer's reference set but with an OTL number provided for each individual signature. Clearly, the results when using the MOTL value are inferior, an outcome that emphasizes the importance of the proposed preprocessing technique.

Table 7 demonstrates the influence of the segmentation profile to the verification performance in case of the CEDAR dataset. Specifically, the following five segmentation scenarios are examined a) using the entire image (EI), b) applying the SP segmentation, c) using only the BRISK keypoints, d) EI in conjunction with SP and finally e) a conjunction of EI, SP and BRISK keypoints. It is evident that the proposed segmentation approach that uses EI, SP and BRISK keypoints outperforms all the other scenarios for all pooling functions (with a minor exception of F$_5$). Moreover, F$_3$ pooling function, and corresponding feature vector persistently provides the best results in all segmentation scenarios and achieves the overall lower error in the proposed segmentation approach (last column of Table 6).

TABLE 6
INFLUENCE (EER %) OF THE VERIFICATION PERFORMANCE WHEN USING A) THE MOTL AND B) THE INDIVIDUAL OTL VALUES

| F(A) | CEDAR | | MCYT | | GPDS300 | | UTSIG | |
|---|---|---|---|---|---|---|---|---|
| | MOTL | Ind.OTL | MOTL | Ind.OTL | MOTL | Ind.OTL | MOTL | Ind.OTL |
| F1 | **2.67** | 3.45 | **3.19** | 4.89 | **2.47** | 6.23 | **7.33** | 12.5 |
| F2 | **4.17** | 5.20 | **8.65** | 10.2 | **3.98** | 8.65 | **11.4** | 24.9 |
| F3 | **1.44** | 2.89 | **2.82** | 4.98 | **1.50** | 5.71 | **6.22** | 10.3 |
| F4 | **7.51** | 9.42 | **16.7** | 20.5 | **15.2** | 20.5 | **20.1** | 26.0 |
| F5 | **3.08** | 5.23 | **3.46** | 5.43 | **3.30** | 7.23 | **7.44** | 13.5 |

TABLE 7
INFLUENCE (EER %) OF THE VERIFICATION PERFORMANCE FOR DIFFERENT SEGMENTATION PROFILES OF THE CEDAR DATASET WITH $L_0$-NORM: NUMBER OF REFERENCE SAMPLES IS 5

| | Entire Image (EI) | SP: $\beta = 2$ | BRISK only | EI & SP | EI & SP &BRISK |
|---|---|---|---|---|---|
| # Features | 60 | 240 | 60 | 300 | 360 |
| F1 | 8.87 | 4.10 | 8.35 | 2.78 | **2.67** |
| F2 | 10.3 | 5.73 | 10.3 | 6.30 | **4.17** |
| F3 | 4.36 | 2.98 | 4.30 | 1.95 | **1.44** |
| F4 | 9.37 | 9.92 | 9.22 | 8.02 | **7.51** |
| F5 | 8.20 | 4.38 | 8.47 | 3.04 | **3.08** |

## TABLE 8
EFFECT OF NUMBER OF ATOMS (K) AND SPARSITY LEVEL ($\rho$) TO THE VERIFICATION ERROR (EER%).

|  | $\beta$ | K=30 | | K=60 | | | K=80 | | |
|---|---|---|---|---|---|---|---|---|---|
|  |  | $\rho=1$ | $\rho=3$ | $\rho=1$ | $\rho=3$ | $\rho=6$ | $\rho=1$ | $\rho=4$ | $\rho=8$ |
| CEDAR | 2 | 1.04 | 1.17 | 0.93 | 0.90 | 1.03 | 0.79 | 0.81 | 0.98 |
|  | 3 | 1.31 | 1.51 | 0.98 | 1.14 | 1.52 | 0.91 | 1.04 | 1.32 |
| MCYT | 2 | 1.85 | 2.42 | 1.55 | 1.63 | 1.72 | 1.49 | 1.53 | 1.78 |
|  | 3 | 1.98 | 2.26 | 1.55 | 1.97 | 1.95 | 1.37 | 1.58 | 1.73 |
| GPDS-300 | 2 | 1.00 | 1.15 | 0.76 | 0.76 | 0.75 | 0.75 | 0.72 | 0.70 |
|  | 3 | 1.07 | 1.20 | 0.93 | 0.92 | 0.92 | 0.90 | 0.85 | 0.86 |

## TABLE 9
RESULTS OBTAINED BY $F_3$ FEATURE AND WITH DIFFERENT TYPES OF NOISE FOR CEDAR DATASET.

| Type of noise | Noise parameters | | | K=60 | | | K=80 | | |
|---|---|---|---|---|---|---|---|---|---|
|  | d | m | v | $\rho=1$ | 3 | 6 | $\rho=1$ | 4 | 8 |
| Salt & pepper | 0.001 | - | - | 1.13 | 1.09 | 1.21 | 0.87 | 0.89 | 1.05 |
|  | 0.01 | - | - | 1.01 | 1.03 | 1.15 | 0.88 | 0.90 | 1.07 |
|  | 0.1 | - | - | 1.03 | 1.07 | 1.21 | 0.90 | 0.92 | 1.22 |
| Gaussian | - | 0 | 0.01 | 1.02 | 1.07 | 1.21 | 0.86 | 0.88 | 1.11 |
|  | - | 0 | 0.1 | 2.94 | 2.94 | 2.98 | 2.86 | 2.91 | 3.21 |
|  | - | 0.2 | 0.01 | 1.02 | 1.04 | 1.10 | 0.86 | 0.94 | 1.13 |

In addition, we report a sensitivity analysis regarding the effect of the verification error that the dictionary learning and SR parameters have. Specifically, we run a set of experiments (KSVD/OMP only) with a variety of parameters: a) the number of dictionary atoms (K= 30 or 60 or 80) and b) the number of sparsity level (1 or 3 for K=30, 1 or 3 or 6 for K=60 and 1 or 4 or 8 for K=80). The experiments resemble cases in which low, medium and high over-completed dictionaries are employed for K=30, 60 and 80, respectively. Similar cases for high, medium and low sparsity levels were also taken into account. This time the experiments were conducted for CEDAR, MCYT and GPDS-300 datasets and with the use of 10 reference samples (12 for GPDS-300) in order to follow up with the most commonly encountered experimental setups in the literature. Table 8 presents the corresponding results. As a first comment on the results, it can be noticed that the use of a "low over-completeness" dictionary and corresponding SR provides lower verification error rates compared to the medium and high overcomplete dictionary /SR cases. Second, it seems that there is a correlation between lower verification error and the increasing number of dictionary atoms. Third, for the CEDAR and MCYT datasets, it is evident that a high sparsity level provides superior results compared to low or medium sparsity level. A possible explanation for this is that a higher order overcomplete dictionary creates and adapts more specific atoms to the learned structures while a higher sparsity level utilizes only the most dominant atom for the representation. This might mislead to the conclusion that a K-means algorithm should be followed instead. However, as stated in Section 2.1, the corresponding alpha coefficient of the most participating atom, during the SR process, is not one, as in the K-means case. Therefore our initial intuition for using SR seems to be justifiable. For the GPDS-300 more atoms are required to participate in order to low the verification error. The higher number of required atoms can be explained by the following facts: a) the GPDS-300 consists of thicker lines, thus more atoms are needed in order to represent a patch and b) GPDS signatures have been acquired with the use of two bounding boxes; so it is anticipated that more atoms should also participate in order to reconstruct the characteristic shapes that exist in a patch.

The performance of the proposed method has been tested also on degraded and noisy data. For this purpose, two different types of noise, namely Salt & Pepper, and Gaussian white noise were employed on the CEDAR dataset with various noise levels. It has been reported that these two types of noise commonly appear in images during the data collection process [53]. Thus, we added noise in the original signature images as follows: a) Salt and Pepper with parameters d=0.001, 0.01 and 0.1 and b) Gaussian with mean and variance parameters i) m=0 & v=0.01, ii) m=0 & v=0.1 and iii) m=0.2 & v=0.01. Given that the images are now corrupted with noise we applied an additional typical preprocessing step with a median 3x3 filter for the case of Salt and Pepper noise. For the Gaussian noise case we also applied a denoising algorithm based on K-SVD which has been used in the literature [54]. The results are provided in Table 9. It can be noticed that the second order statistic feature ($F_3$) still outperforms any other type of feature. In addition, the overall performance of the proposed method remains relatively stable in the presence of noise. A closer insight reveals also that the part of the feature extraction method that is affected mostly from noise is the equimass segmentation and the corresponding spatial pyramid pooling.

As an additional comment on the results we could state that, as usual, the method does not perform well when the signatures of a signatory are either too simple or/and they present a significant amount of instability. This effect is quite strong in the case of using a limited number of reference samples (e.g. 5). Although the addition of more signature samples alleviates this effect, we report that this is still an open issue that a SV system must cope with. Perhaps a disadvantage of the used signature datasets so far is that there is not some additional information (e.g. the characterization by the signatories themselves of some of the genuine samples as outliers) which would allow us to tune better the learning stage of the classifier.

Perhaps the most challenging task on SV is the comparison of the results emanating from several state-of-the-art systems. This turns to become a difficult issue given the fact that, traditionally, the efficiency of SV's is evaluated under two independent scenarios: random forgeries and skilled forgeries. It has been reported [52] that while these scenarios are not necessarily unfitting, their use may lead to a misinterpretation of the results. Additionally, this kind of has been also stated to be unrealistic in typical real world cases [52]. Tables 10-12 outline and compare the $EER^S_{User\_Threshold}$ based results of the proposed method, with emphasis to the F3 pooling, to a number of state-of-the-art SV related methods regarding the four utilized datasets, on an EER or average error (AER) basis. The results clearly demonstrate that the proposed method achieves low verification errors that are at least comparable to the ones derived from other state-of-the-art meth-ods. Having in mind the diversity of all other methods we comment that with respect to CEDAR, MCYT and UTSIG datasets, the reported error rate outperforms almost all other methods. Regarding to the GPDS case the reported results are also quite low, overpassed only by the method reported in [8] which is a DL oriented method.

## 6. CONCLUSIONS

In this paper the potential of Sparse Representation on creating discriminative features for accurate and efficient offline signature verification is presented. We thoroughly investigated the major aspects on the selection of the appropriate SR approach, and examined the effects of the associated parameters like the number of atoms, the sparsity level and the regularization function. We demonstrated that approximate greedy techniques can deploy the full potential of SR in a SV system, in a computationally attractive manner. We described a novel pooling scheme tailored to the problem of SV, and evinced that $2^{nd}$ order statistics - standard deviation - can form more discriminative pooling functions in cases where signals exhibit degenerate structures. Finally, we proposed a carefully designed system, encompassing a novel algorithm for the automatic selection of the optimal thinning level, which is able to harness the power of SR in order to create a discriminative signature descriptor which obtains state-of-the-art results on the most challenging signature datasets. The method seems also to handle relative well noisy images that have undergone a typical preprocessing denoising step.

TABLE 10
COMPARISONS WITH SV STATE-OF-THE-ART SV TECHNIQUES FOR CEDAR DATASET.

| 1st author Ref # | Method | # Ref. | AER / EER |
|---|---|---|---|
| Kumar R. [55] | Signature Morphology (WI) | 1 | 11.6 / 11.8 |
| Kumar R [56] | Surroundness (WI) | 1 | 8.33 |
| Kumar M. [57] | Chord moments | 16 | 6.02 |
| Chen [58] | Gradient+concavity | 16 | 7.90 |
| Chen [58] | Zernike moments | 16 | 16.4 |
| Kalera [46] | G. S & C (WI) | 16 | 21.9 |
| Zois [7] | Partially Ordered Sets | 5 | 4.12 |
| Guerbai [59] | Curvelet Transform (WI) | 12 | 5.60 |
| Serdouk [60] | Gradient LBP + LRF | 16 | 3.52 |
| Hafemann [29] | SigNet / SigNet-F | 12 /12 | 4.76 / 4.63 |
| Hafemann [8] | SigNet-SPP-300dpi | 10 | 3.60 |
| Hafemann [8] | SigNet-SPP-300dpi (fine-tuned) | 10 | 2.33 |
| Bharathi [61] | Chain code | 12 | 7.84 |
| Okawa [12] | B.O.W with KAZE | 16 | 1.60 |
| Okawa [43] | V.L.A.D with KAZE | 16 | 1.00 |
| Ganapathi [62] | Gradient Direction | 14 | 6.01 |
| Shekar [63] | Morphological Pattern Spectrum | 16 | 9.58 |
| Hamadene [64] | Directional Co-occurrence (WI) | 5 | 2.11 |
| Zois [20] | Archetypes | 5 | 2.07 |
| Zois [65] | Tree structured sparsity | 5 | 2.30 |
| Dutta [13] | Compact Correlated Features | N/A | 0.00 |
| Tsourounis [66] | Deep Sparse Coding | 5 | 2.82 |
| **Proposed** | **K-SVD/OMP** | **10** | **0.79** |

TABLE 11
COMPARISONS WITH SV STATE-OF-THE-ART TECHNIQUES FOR MCYT-75 DATASET.

| 1st author Ref # | Method | # Ref. | AER / EER |
|---|---|---|---|
| Vargas [48] | L.B.P | 5 / 10 | 11.9 / 7.08 |
| Zois [7] | Partially Ordered Sets | 5 | 6.02 |
| Gilperez [67] | Contours | 5 / 10 | 10.18 / 6.44 |
| Alonso-Fernadez [68] | Slant and Envelope | 5 | 22.4 |
| Fierrez-Aguilar [69] | Global and Local Slant | 10 | 9.28 |
| Wen [70] | Invariant Ring Peripheral | 5 | 15.0 |
| Ooi [71] | Discrete Radon Transform | 5 / 10 | 13.86 / 9.87 |
| Soleimani [9] | HOG & Deep-MML | 5 / 10 | 13.44 / 9.86 |
| Hafemann [29] | SigNet / SigNet-F | 10 / 10 | 2.87 / 3.00 |
| Hafemann [8] | SigNet-SPP-300dpi | 10 | 3.60 |
| Hafemann [8] | SigNet-SPP-300dpi (finetuned) | 10 | 2.33 |
| Serdouk [14] | H.O.T | 10 | 18.15 |
| Okawa [43] | BoVW - VLAD - KAZE | 10 | 6.4 |
| Zois [20] | Archetypes | 5 | 3.97 |
| Zois [65] | Tree structured sparsity | 5 | 3.52 |
| **Proposed** | **K-SVD/OMP ($F_3$)** | **10** | **1.37** |

TABLE 12
COMPARISONS WITH SV STATE-OF-THE-ART SV TECHNIQUES FOR GPDS AND UTSIG DATASETS.

| Dataset | 1st author Ref # | Method | # Ref. | AER / EER |
|---|---|---|---|---|
| GPDS 100 | Serdouk [60] | Gradient LBP + LRF | 16 | 12.52 |
| | Vargas [48] | L.B.P | 12 | 6.20 |
| | Favorskaya [72] | Global traits | 15 | 12.72 |
| | Bharathi [61] | Chain code | 12 | 9.64 |
| 150 | Hu [73] | LBP & HOG & GLCM | 10 | 7.66 |
| 160 | Yılmaz [74] | HOG-LBP & SVM | 12 | 6.97 |
| | Nguyen [75] | MDF, Energy, Maxima | 12 | 17.25 |
| | Guerbai [59] | Curvelet Transform | 8 | 15.95 |
| | Ferrer [76] | Geometric Features | 16 | 9.64 |
| | Alaei [53] | LBP based | 8 / 12 | 13.85 / 11.74 |
| 300 | Zois [7] | Partially Ordered Sets | 5 | 5.48 |
| | Kumar R. [56] | Surroundness | 1 | 13.76 |
| | Eskander [77] | ESC-DPDF | 12 | 17.82 |
| | Soleimani [9] | HOG & Deep-MML | 10 | 20.94 |
| | Hafemann [29] | SigNet / SigNet-F | 12 /12 | 3.15 / 1.69 |
| | Hafemann [8] | SigNet-SPP-300dpi | 10 | 3.15 |
| | Hafemann [8] | SigNet-SPP-300dpi-F | 10 | 0.41 |
| | Parodi [78] | Circular Grid | 13 | 4.21 |
| | Serdouk [14] | H.O.T | 10 | 8.76 / 9.30 |
| | Pirlo [79] | Cosine similarity | 12 | 7.20 |
| | Pirlo [80] | Optical flow | 6 | 4.60 |
| | Hamadene [64] | Directional Co-occurrence | 5 | 18.42 |
| | Dutta [13] | Compact Correlated Features | N/A | 11.21 |
| 960 | Yilmaz [81] | Two channel CNN (WI) | 12 | 2.05 |
| | | Two channel CNN & Signet-F | 12 | 0.88 |
| 300 | **Proposed** | **K-SVD/OMP ($F_3$)** | **12** | **0.70** |
| UTSIG | Soleimani [9] | HOG-DMML | 12 | 17.6 |
| | Soleimani [50] | Fixed-point arithmetic | 12 | 29.7 |
| | **Proposed** | **K-SVD/OMP ($F_3$)** | **12** | **6.22** |

Our future research plans include the exploitation of SR related techniques in order to construct a universal dictionary with the use of samples originating from a wide and diverse set of persons (instead of just only one signatory) in order to develop an efficient writer independent signature verifier.

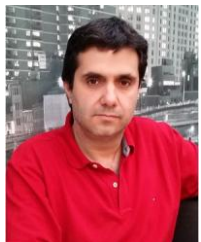

**Elias N. Zois** received his Bachelor's degree in Physics from the University of Patras (UoP), Patras, Greece (1994), the M.Sc. degree in Electronic Engineering from UoP (1996), along with his Ph.D. from UoP (2000). From 2000 to 2008, he has been working as an Adjunct Professor at the Technological and Educational Institute of Athens. Currently he is an Assistant Professor at the University of West Attica. His research interests include among other, computer vision, image processing, machine learning, and biometrics.

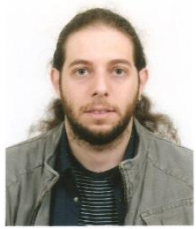

**Dimitrios Tsourounis** received his B.Sc. in Physics and M.Sc. in Electronics and Information Processing from the Physics Dept., University of Patras, Greece in 2015 and 2017. He is now a Ph.D. candidate in Machine Learning at the Physics Department of the University of Patras. His research interests include Machine Learning, Pattern Recognition, Image Processing, Computer Vision applications and Biometrics.

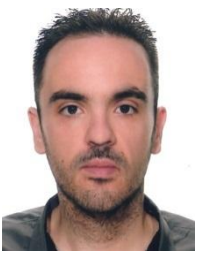

**Ilias Theodorakopoulos** received a B.Sc. degree in Physics in 2007, a M.Sc. degree in Electronics and Computers in 2009 and a Ph.D degree in Pattern Recognition & Manifold Learning in 2014, all from University of Patras. He works as a data scientist and researcher in the fields of Machine Learning and Data Analysis, having more than 20 publications and 3 patents (granted and pending). His research interests include theoretical foundations of Machine Learning, applications on Computer Vision, Biometric Identification and Biomedical Data Analysis**.**

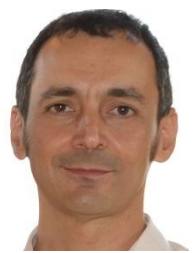

**Anastasios L. Kesidis** received his B.S and Ph.D degrees in Electrical and Computer Engineering from Democritus University of Thrace, Xanthi, Greece, in 1995 and 2001, respectively. From 2000 to 2002 he worked as a postdoctoral Research Fellow in Centre for Vision, Speech and Signal Processing, University of Surrey, UK and from 2005 to 2010 he joined the National Centre for Scientific Research "Demokritos", Athens, Greece as a Research Scientist. Currently he is an Associate Professor in the Department of Surveying and Geoinformatics Engineering at the University of West Attika, Greece. His research interests include image processing, pattern recognition and machine learning.

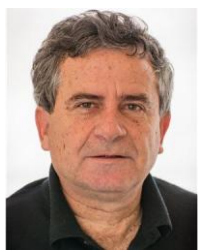

**George Economou** received his Bachelor's degree in Physics from the University of Patras (UoP), Patras, Greece (1976), the M.Sc. degree in Microwaves and Modern Optics from University College, London, U.K. (1978), and the Ph.D. degree in 'Fiber optic sensor systems' from UoP (1989). From 1983 to 1985, he has been working as a Visitor Research Assistant at University College London. He is currently Professor with the Department of Physics University of Patras. His research interests include computer vision, image processing, machine learning, pattern recognition and biometrics.

## APPENDIX A – SPARSE CODING

Informally, the objective of Sparse Representation (SR) is to encode a set of signals as a linear combination of a few elements of a predefined set (dictionary) whose elements are defined as basis vectors or atoms. The sparse dictionary is usually overcomplete i.e. the number of dictionary elements is higher than the input signal's dimensions. Several techniques for SR have been successfully applied in the fields of computer vision, pattern recognition and machine learning. The key idea is to represent an observed phenomenon through the activation of only as few components as possible. It has been shown that the $V_1$ part of the brain which receives visual stimuli, is alleged to perform a similar process with sparsity objectives [1]. For an overcomplete dictionary $\mathbf{D} = [\mathbf{d}^1, \mathbf{d}^2, ..., \mathbf{d}^K] \in \mathbb{R}^{n \times K}$ with $K > n$ and for M input signals $\mathbf{X} = [\mathbf{x}^1, \mathbf{x}^2, ... \mathbf{x}^M] \in \mathbb{R}^{n \times M}$, the formulation of SR is usually expressed with the following equivalent forms of regularized or constraint optimization problem expressed by (1) and (2) respectively:

$$\frac{1}{M} \sum_{i=1}^{M} \min_{\boldsymbol{\alpha}^i} (\frac{1}{2} \|\mathbf{x}^i - \mathbf{D}\boldsymbol{\alpha}^i\|_2^2 + \lambda \psi(\boldsymbol{\alpha}^i)) \quad (1)$$

$$\frac{1}{M} \sum_{i=1}^{M} \min_{\boldsymbol{\alpha}^i} (\frac{1}{2} \|\mathbf{x}^i - \mathbf{D}\boldsymbol{\alpha}^i\|_2^2), \text{ s.t: } \psi(\boldsymbol{\alpha}^i) \leq \rho, \forall_{i=1:M} \quad (2)$$

In the above expressions (1), (2), the matrix $\mathbf{A} = [\boldsymbol{\alpha}^1, \boldsymbol{\alpha}^2, ... \boldsymbol{\alpha}^M] \in R^{K \times M}$ represents the sparse coefficients for the M input signals, $\|\mathbf{x}^i - \mathbf{D}\boldsymbol{\alpha}^i\|_2^2$ is the reconstruction error, $\lambda$ is the regularizer parameter or Lagrange multiplier and $\psi(\cdot)$ is the sparsity-inducing term [2]. The embedded term $\psi(\cdot)$ is defined to be the $l_p$-norm (defined for $1 \leq p \leq \infty$) of the coefficients $\boldsymbol{\alpha}^i$, i.e. $\psi(\boldsymbol{\alpha}^i) = l_p(\boldsymbol{\alpha}^i) = \|\boldsymbol{\alpha}^i\|_p = \left(\sum_{j=1}^{K} (\boldsymbol{\alpha}^i[j])^p\right)^{1/p}$ for a specific value of p. The most popular forms are the ones that rely on the $l_0$-norm and $l_1$-norm, i.e. for p=0 or p=1. Specifically, the $l_0$-norm or pseudo-norm is equal to the count of non-zero elements $\rho$ (sparsity level) of the representation vector $a^i$ and so it is the most direct sparsity measure. However, this selection leads to a combinatorial NP-hard optimization problem, whose solution can only be approximated [3]. On the contrary, the $l_1$-norm leads into a convex relaxation of the coding problem and has been proved that encourages sparse solutions [3]. Thus, the sparsity inducing term is usually utilized by the $l_0$-norm which is a greedy non-convex approximation and the $l_1$-norm which is a convex relaxation named Lasso [4].

To obtain a satisfactory solution of Sparse Coding with the use of the $l_0$-norm regularization term, greedy algorithms can be utilized in order to seek and provide an approximate optimized solution. The idea behind the greedy strategy is to always seek for the atom with the strongest relation to the sample under examination, in an effort to aggressively reduce the reconstruction error in the least-squares sense. The orthogonal matching pursuit (OMP) [5] along with a number of variations like regularized OMP, stage wise OMP, sparsity adaptive matching pursuit, are typical representatives of greedy algorithms. Given a dictionary $\mathbf{D}$ and any sample $\mathbf{x}^i$, OMP sequentially selects the atoms with the highest correlation to the respective sample's residual. At a step $s: 0 < s \leq \rho$ the selected atom is given by: $k_s = \arg\max |(\mathbf{d}^j)^T \mathbf{r}_{s-1}|$, where $\mathbf{r}_{s-1}$ is the current residual. Once an atom is selected, the $\mathbf{x}^i$ signal is projected onto the span of currently selected atoms as: $\hat{\mathbf{a}}_s = (\mathbf{D}_{V_s})^+ \mathbf{x}^i$, where $V_s = V_{s-1} \cup k_s$ is the set of indices pointing at the currently selected dictionary atoms, and $\mathbf{D}_{V_s}$ is the subset of dictionary indexed by $V_s$. The new residual is now given by $\mathbf{r}_s = \mathbf{x} - \mathbf{D}_{V_s} \hat{\mathbf{a}}_s$ while the process now repeats until either $\rho$ atoms are selected or the residual magnitude minimizes. In this work we utilized the batch-OMP implementation [6], which makes use of Cholesky decomposition in order to reduce the computational cost of repeated re-projections, as a means to assess the efficiency of greedy SR methods in signature verification.

The use of the $l_1$-norm, has been also broadly proposed for sparse solutions since it provides an analytical solution and can be solved in polynomial time. Solvers for handling the $l_1$-norm oriented SR problems, include the well-known basis pursuit [7] and the Lasso [4] among others. A number of methods for solving the $l_1$-norm problems rely on coordinate descent methods, which have been found to be efficient enough in cases where dictionary atoms exhibit low correlation. Unlike, when the atoms of the dictionary are highly correlated, homotopy-based methods can be applied. The homotopy methods solve the regularized sparse coding optimization problem, as expressed by (1), and track the entire regularization path (i.e. the solutions for all possible values of $\lambda$), like the least angle regression (LARS-Lasso) algorithm [8]. The LARS-Lasso algorithm calculates the solution path by repeatedly decreasing the value of $\lambda$ and using as a warm-restart of the previously calculated solution. The uniqueness of the solution for a specific value of the parameter $\lambda$, i.e. the existence of a single normalized path solution, is ensured and it can be proved that the solution path is piecewise linear [9]. This property is very important since the algorithm follows the direction of each segment until it reaches a critical point, i.e. where either a non-zero element becomes zero (so it is removed from the active set of coefficients) or a new non-zero element is added to the active set of coefficients. Therefore, the homotopy method initializes with an empty set of coefficients, and iteratively updates it by one variable at a time. The complexity of the method relies in reversing the covariance matrix of the selected atoms at each critical point in order to update the active set of coefficients, which is performed by the Cholesky decomposition or the Woodbury formula. In this work the LARS-Lasso algorithm, which is a part of the SPAMS toolbox [9], has been utilized to investigate the efficiency of sparse coding with convex relaxation.

## APPENDIX B – DICTIONARY LEARNING

The most efficient way for dictionary construction is through a learning process that enables the dictionary (i.e. atoms) to be fitted to the input-training data. The most common dictionary learning methods are unsupervised and have been mainly utilized for image processing problems, such as image compression and super resolution.

Although these methods are not explicitly enforcing discriminative behavior to the sparse coefficients since the cost function relies only on reconstruction error, they have been used for classification tasks with remarkable results [10]. The sparse representation problem (i.e. dictionary learning and sparse coding) is a joint optimization problem with respect to dictionary $\mathbf{D}$ as well as the coefficients $\mathbf{A} = \{\boldsymbol{\alpha}^i\}$ and it is expressed as follows:

$$\min_{\mathbf{D} \in C, \mathbf{A} \in R^{K \times M}} (\frac{1}{2}\|\mathbf{X} - \mathbf{DA}\|_F^2 + \lambda \|\boldsymbol{\alpha}^i\|_p) \quad (3)$$

$$\min_{\mathbf{D} \in C, \mathbf{A} \in R^{K \times M}} (\frac{1}{2}\|\mathbf{X} - \mathbf{DA}\|_F^2) \text{ s.t: } \|\mathbf{A}\|_p \leq \rho \quad (4)$$

where the set $C$ of the dictionary atoms is usually defined to be the convex set of matrices that satisfy the following constraint: $C \triangleq \{\mathbf{D} \in R^{n \times K} \text{ s.t: } \forall_{j=1:M}, (\mathbf{d}^j)^T \mathbf{d}^j \leq 1\}$ in order to avoid large values, which consequently might lead to arbitrarily small values for the coefficients.

The greedy approximation of $l_0$-norm regularized SR is addressed in this work by means of the KSVD [11, 12] and the associated OMP algorithm for dictionary learning and sparse coding respectively. For the convex relaxation approach, the online learning method [9] along with the LARS-Lasso algorithm have been employed for the corresponding tasks. In the case of the KSVD/OMP algorithm pair, the notation of eq. (4) has been selected due to the fact that the use of the $l_0$ sparsity constraint $\|\mathbf{A}\|_0 \leq \rho$ with $\rho \in \mathbb{N}^+$, intuitively assigns as a design parameter the integer number of the non-zero elements $\rho$. On the other hand, for the case of online/LARS-Lasso algorithm pair the notation of eq. (3) has been followed due to the fact that the interpretation of the parameter $\rho_1 \in \mathbb{R}$ in the $\ell_1$ constraint $\|\mathbf{A}\|_1 \leq \rho_1$ does not provide any thoughtful intuition regarding the sparsity of the atoms as a design parameter. Instead, the parameter $\lambda$ of eq. (3) shall be used as the design parameter of the method which now controls the sparsity. From a theoretical point of view, for an appropriate dictionary, the solution obtained with the use of the $\ell_1$-norm is equivalent to the one provided by another $\ell_0$-norm based solution with full probability [13, 14], [3]. However, this $\ell_1 / \ell_0$ equivalence phenomenon, does not hold in the case of the joint regularized and constrained dictionary learning problem since it has been reported that different formulations of the dictionary learning problem do not admit the same solutions even for the case of using the same $\ell_1$-norm [9]. That is the reason behind our decision to evaluate the performance of both $\ell_0$ and $\ell_1$-norm approaches.

# Appendix C – Main Code - Pseudocode of the proposed method

**Main Algorithm:** Dictionary Learning and sparse coding of signature images
**Require:** The signatures of a dataset,
**BEGIN**
1: **SELECT:** One signatory (*wrn*) from the signature dataset.
2: **DEFINE:** *Learning* SET from the *wrn*-signatory

   $\{\omega^{\oplus}\}$: #$N_{G\text{-REF}}$ Genuine Reference Samples, randomly selected (e.g. 5)

   $\{\omega^{-}\}$: #$2 \times N_{G\text{-REF}}$ Genuine Samples from other signatories (e.g. 10)
   $PS = patch\_size$ (e.g. 5); $PS^2$: $PS \times PS$
   $p$: *number of dictionary atoms* (> 2×patch_size, e.g. p=60),
3. **DEFINE:** *Testing Set* for the *wrn*-signatory
   Remaining Genuine Samples (e.g. 19 for CEDAR)
   All forgeries (e.g. 24 for CEDAR)
   Random forgeries (e.g. 54 for CEDAR)
4: **CALL <u>ALGORITHM 1</u>** and **DEFINE** MOTL for $\omega^{\oplus}$ samples
5: **FOR** j = 1 **UP TO** size of $\{\omega^{\oplus}\}$
6:   **PREPROCESS:** Threshold, *thin_with_MOTL* the signature image.
7:   **EXTRACT** all patches $X \in \mathbb{R}^{PS^2 \times (\#patches)}$; **CENTER** each patch $X_k \in \mathbb{R}^{PS^2 \times 1}$
8:   **IF** j=1 then
       **USE** Dictionary Learning algorithm (e.g. K-SVD). **RETURN** $D \in \mathbb{R}^{PS^2 \times p}$
     **ELSE**
       **UPDATE** $D \in \mathbb{R}^{PS^2 \times p}$ with new patch matrix $X \in \mathbb{R}^{PS^2 \times (\#patches)}$
     end_IF
9: end_fOR
10: **FOR any** signature of $\{\omega^{\oplus}, \omega^{-}, Testing\ Set\}$: **APPLY** steps 6, 7:
11:   **USE** OMP or LASSO for evaluation of sparse coefficients $A \in \mathbb{R}^{p \times (\#patches)}$
12:   **USE** some poolingFunction( $A \in \mathbb{R}^{p \times (\#patches)}$ ), **CREATE** features $F \in \mathbb{R}^{p \times 1}$
13:     **FOR** any other signature segment (Equimass, BRISK)
14:       **REPEAT** steps 6, 7, **APPEND** and **UPDATE** $F$ to $F \in \mathbb{R}^{[p \times (\#segments)] \times 1}$
15:     end_FOR
16: end_FOR. % *Features for all signatures*
17: **TRAIN** and **CROSS_VALIDATE** with $\{\omega^{\oplus}, \omega^{-}\}$
18: **TEST** with *Testing Set*

# Appendix D – OTL (and MOTL) Pseudocode

**Algorithm 1:** Define optimal thinning level.
**Require:** $B_i$, i=1: $N_{G-REF}$ Reference binary set of images, $N_p$- patch size.
1: **for** i=1: $N_{G-REF}$
   2: **set th_lev EQUAL TO 0**    % (thinning level)
   3:   **Repeat**
   4:     $TH^i_{th\_lev}$ =THIN_OPERATION($B_i$, **th_lev**)
   5:     **Create** map with locations of $TH^i_{th\_lev}$ signature pixels.
   6:     **for** j = each signature pixel of the map
         Impose a $N_p \times N_p$ patch, centered at the coordinates of the j-pixel.
         **Measure** the number of signature pixels that reside in the patch.
         **Normalize** by $N_p \times N_p$ and **ASSIGN** value to LD(j).
         end for
   7:   **Evaluate** mean value of LD and **ASSIGN** value to PD(**th_lev**).
   8:   **th_lev** ← **th_lev** +1            (**increase thinning level**)
   9:   **Until** $TH^i_{th\_lev} = TH^i_{th\_lev+1}$    (has idempotent stage reached?)
   10.  **Find** minimum of the difference of the PD function,
   11.    **RETURN** minimum index to **OTL(i)**.
12: **end for**
**Return: MOTL:** The median value of **OTL**.